\documentclass[conference]{IEEEtran}
\IEEEoverridecommandlockouts
\usepackage{cite}
\usepackage{amsmath,amssymb,amsfonts}
\usepackage{algorithmic}
\usepackage{textcomp}
\usepackage{xcolor}
\usepackage{url}
\usepackage{algorithm}
\usepackage{graphicx}
\usepackage{amsmath}
\usepackage{graphicx}  
\usepackage{subfig}    
\usepackage{caption}  
\usepackage{color}
\usepackage{framed}
\usepackage{comment}
\usepackage{tcolorbox}
\usepackage{makecell}
\usepackage{subcaption}
\usepackage{subfloat} 
\usepackage{amsthm}
\usepackage{multirow}
\usepackage{multicol}
\newtheorem{lemma}{Lemma}

\tcbuselibrary{skins,breakable}

\def\BibTeX{{\rm B\kern-.05em{\sc i\kern-.025em b}\kern-.08em
    T\kern-.1667em\lower.7ex\hbox{E}\kern-.125emX}}
\begin{document}

\title{GoPrune: Accelerated Structured Pruning \\with $\ell_{2,p}$-Norm Optimization\thanks{This work was supported by the National Natural Science Foundation of China under Grant 12371306. \textit{(Corresponding author: Xianchao Xiu.)}}}

\author{\IEEEauthorblockN{Li Xu}
\IEEEauthorblockA{\textit{School of Mechatronic Engineering and Automation} \\
\textit{Shanghai University}\\
Shanghai, China \\
xljackson@shu.edu.cn}
\and
\IEEEauthorblockN{Xianchao Xiu}
\IEEEauthorblockA{\textit{School of Mechatronic Engineering and Automation} \\
\textit{Shanghai University}\\
Shanghai, China \\
xcxiu@shu.edu.cn}
}

\maketitle

\begin{abstract}
Convolutional neural networks (CNNs) suffer from rapidly increasing storage and computational costs as their depth grows, which severely hinders their deployment on resource-constrained edge devices. Pruning is a practical approach for network compression, among which structured pruning is the most effective for inference acceleration. Although existing work has applied the $\ell_p$-norm to pruning, it only considers unstructured pruning with $p\in (0, 1)$ and has low computational efficiency. To overcome these limitations, we propose an accelerated structured pruning method called GoPrune. Our method employs the $\ell_{2,p}$-norm for sparse network learning, where the value of $p$ is extended to $[0, 1)$. Moreover, we develop an efficient optimization algorithm based on the proximal alternating minimization (PAM), and the resulting subproblems enjoy closed-form solutions, thus improving compression efficiency. Experiments on the CIFAR datasets using ResNet and VGG models demonstrate the superior performance of the proposed method in network pruning. Our code is available at \url{https://github.com/xianchaoxiu/GoPrune}.
\end{abstract}

\begin{IEEEkeywords}
Convolutional neural networks, pruning, $\ell_{2,p}$-norm, proximal alternating minimization, sparse optimization
\end{IEEEkeywords}

\section{Introduction}

Convolutional neural networks (CNNs) have achieved remarkable success in various computer vision tasks \cite{li2021survey}, including image classification \cite{xi2025mctgcl}, object detection \cite{liu2025lightweight}, and semantic segmentation \cite{chen2025weakly}. However, the ever-increasing depth and width of modern CNN architectures have led to substantial computational costs and memory footprints, posing significant challenges for deployment on edge devices such as mobile phones, drones, and robots \cite{chen2024review}.

During the past few decades, numerous researchers have focused on network compression, including knowledge distillation \cite{moslemi2024survey}, quantization \cite{rokh2023comprehensive}, and pruning \cite{cheng2024survey}. Compared to the former two techniques,  network pruning has gained significant attention due to its simplicity and ease of implementation. Unlike unstructured pruning, structured pruning \cite{he2023structured} directly discards entire channels, filters, or layers, yielding thinner models that are compatible with off-the-shelf deep learning libraries and general-purpose processors, thereby being more hardware-friendly. Traditional structured pruning methods typically follow a three-stage pipeline: (1) training a full-capacity model, (2) evaluating the importance of layer structures, (3) pruning less-important structures, followed by fine-tuning.

It is worth noting that both structured and unstructured pruning methods revolve around regularization to identify sparse sub-networks. For example, Kumar et al. \cite{kumar2021pruning} adopted the $\ell_{1}$-norm to prune filters and neurons in CNNs, while Kang et al. \cite{kang2022pruning} employed the $\ell_{1/2}$-norm for individual weights of network pruning and proved its convergence.  Recently, Gao et al. \cite{gao2023structural} applied the $\ell_{2}$-norm to the selected sub-network and pruned redundant channels in CNNs.
Lee et al. \cite{lee2024pruning} utilized the nuclear norm to prune redundant channels in the network. Generally speaking, current regularization-based pruning methods mainly focus on the $\ell_{1}$-norm or $\ell_{2}$-norm, while the $\ell_{0}$-norm \cite{de2024compression} is rarely explored. In fact, the $\ell_{0}$-norm counts the number of non-zero elements, thereby providing a more effective balance between sparsity and performance. Nevertheless, it is an NP-hard problem. Very recently, Ji et al. \cite{ji2025network} proposed the $\ell_{p}$-norm and developed an optimization algorithm based on alternating direction method of multipliers (ADMM). Unfortunately, it is an unstructured pruning method and has two shortcomings. On the one hand, the $\ell_{p}$-norm can only take values within the range of $(0,1)$, excluding the $\ell_{0}$-norm, which limits the performance \cite{liu2024towards}. On the other hand, the resulting subproblems require iterative computation, leading to a large computational load. \textit{A natural question then arises: can this method be extended to structural pruning and solve both shortcomings simultaneously?}

In this paper, we propose an accelerated structured pruning framework called  GoPrune. Specifically, GoPrune integrates the $\ell_{2,p}$-norm with $p\in [0,1)$ into the training objective to induce filter-level sparsity during optimization, enabling the model to learn the importance of each channel automatically. After training, the one-shot channel pruning is performed based on the learned channel importance, followed by a lightweight fine-tuning stage to recover accuracy. Overall, the main contributions can be summarized as follows.
\begin{itemize}
    \item We propose a novel structured pruning strategy by integrating the $\ell_{2,p}$-norm with $p\in [0,1)$, which covers $\ell_{p}$-norm with $p\in (0,1)$ in \cite{ji2025network}  as a special case.
    \item We develop an efficient algorithm with closed-form solutions, which significantly reduces the compression time.
    \item We verify its superiority through numerical experiments on the CIFAR datasets.
\end{itemize}

\section{Methodology}

\subsection{Notations}
In this paper, weight tensors are represented by capital letters. For any $X\in\mathbb{R}^{C_{in}\times C_{out}\times k\times k}$, the $j$-th channel is denoted by $X_{:j::} \in \mathbb{R}^{C_{in}\times k\times k}$ with the corresponding vectorization being $\textrm{vec}(X_{:j::})$. The Frobenius norm is defined as \begin{equation}
    \|X\|_{\textrm{F}} = (\sum_{i=1}^{C_{in}} \sum_{j=1}^{C_{out}} \sum_{m=1}^{k} \sum_{n=1}^{k}  X_{ijmn}^2)^{1/2},
\end{equation} 
where $X_{ijmn}$ is the $ijmn$-th element.
For $p\in (0,1)$, the $\ell_{2,p}$-norm is defined as \begin{equation}
    \|X\|_{2,p}=(\sum_{j=1}^{C_{out}}\|\textrm{vec}(X_{:j::})\|^p)^{1/p}. 
\end{equation}
In addition, $\|X\|_{2,0}$ counts the number of non-zero channels. A further notation will
be introduced wherever it appears.

\subsection{Model Construction}
As stated in \cite{ji2025network}, different from the $\ell_1$-norm and $\ell_2$-norm, the $\ell_p$-norm with $p\in (0,1)$ can achieve better sparsity and is closer to the $\ell_0$-norm. Assume that $W\in \mathbb{R}^{C_{in} \times C_{out} \times k \times k}$ represents the weights of the neural network and is a four-dimensional tensor. Thus, the loss function is defined as
\begin{equation}\label{pp}
    \min_W\quad L(W)+\lambda\|W\|_p^p,
\end{equation}
where $p\in (0,1)$ and $\lambda$ is the parameter to control the sparsity. $L(W)=l(W)+\alpha\|W\|_\textrm{F}^2$, of which $l(W)$ is the cross-entropy loss function and $\alpha$ is the decay weight to prevent overfitting.

Motivated by \cite{xiu2025bi}, we leverage the $\ell_{2,p}$-norm for network pruning and construct
   \begin{equation}\label{O-function}
       \min_{W}\quad L(W) + \lambda\|W\|_{2,p}^p,
   \end{equation}
where $p\in [0,1)$. Obviously, it can be regarded as a general case of \eqref{pp}.

\subsection{Optimization}
In the following, an efficient optimization algorithm is provided based on the proximal alternating minimization (PAM) technique. By introducing the auxiliary variable $W=U$, \eqref{O-function} can be rewritten as  
   \begin{equation}\label{mod-1}
   \begin{aligned}
              \min_{W,U}\quad &L(W) + \lambda\|U\|_{2,p}^p\\
              \textrm{s.t.}\quad & W=U.
   \end{aligned}
   \end{equation}
The unconstrained version of \eqref{mod-1} is
\begin{equation}\label{Lagrangian}
     \min_{W,U}\quad L(W) + \lambda\|U\|_{2,p}^{p} + \frac{\beta}{2}\|W -U\|_\textrm{F}^2,
\end{equation}
where $\beta>0$ is the penalty parameter. Denote the objective function of \eqref{Lagrangian} as $f(W, U)$. Under the PAM framework, each variable can be updated alternately via 
\begin{equation}\label{L_k}
\begin{cases}
W^{k+1} \in \min\limits_{W}~ f(W,U^k)+\frac{\rho_1}{2}\|W-W^k\|_\textrm{F}^2,\\
    U^{k+1}  \in \min\limits_{U}~ f(W^{k+1},U)+\frac{\rho_2}{2}\|U-U^k\|_\textrm{F}^2,
\end{cases}
\end{equation}
where $\rho_1, \rho_2 >0$ and $k$ is the iteration number.
The overall scheme is presented in Algorithm \ref{algorithm}, and the update rules for $W$ and $U$ are analyzed as follows.

\begin{algorithm}[t]
    \caption{Optimization algorithm}\label{algorithm}
    \textbf{Input:} $W \in\mathbb{R}^{C_{in}\times C_{out}\times k\times k}$,  $p$, $\alpha$, $\lambda$, $\beta$, $\rho_1$, $\rho_2$, $\eta$\\
    \textbf{Initialize:} $k=0$\\
    \textbf{While} not converged \textbf{do}
        \begin{algorithmic}[1]
            \STATE Update $W^{k+1}$ by (\ref{L_k})
            \STATE Update $U^{k+1}$ by (\ref{L_U})
            \STATE Check convergence: If  $\textrm{epoch} \geq 15$, then stop; otherwise $k=k+1$
        \end{algorithmic}
    \textbf{End While}\\
    \textbf{Output:}  $W^{k+1}$, $U^{k+1}$
\end{algorithm}

\begin{algorithm}[t]
    \caption{Solution of (\ref{L_k})}\label{SGDalgo}
    \textbf{Input:}  $W^{k}, U^k \in\mathbb{R}^{C_{in}\times C_{out}\times k\times k}$, $\eta$, $\beta$, $\rho_1$\\
    \textbf{Initialize:} $t=0$\\
    \textbf{While} not converged \textbf{do}
        \begin{algorithmic}[1]
            \STATE Update $W^{k,t}$ by (\ref{SGD})
            \STATE Check convergence: If  $t \geq T_{\max} $, then stop; otherwise $t=t+1$
        \end{algorithmic}
    \textbf{End While}\\
    \textbf{Output:}  $W^{k+1}$
\end{algorithm}

\subsubsection{Update $W$} The subproblem can be expressed as
\begin{equation}\label{W}
\begin{aligned}
\min_W \quad L(W)
    +\frac{\beta}{2}\|W-U^k\|_\textrm{F}^2
    +\frac{\rho_1}{2}\|W-W^k\|_\textrm{F}^2 .
\end{aligned}
\end{equation}
By expanding the norm terms in (\ref{W}), it can
be rewritten in the form of
\begin{equation}
    \min_W \quad L(W)
    +\frac{\beta+\rho_1}{2}\|W-M\|_\textrm{F}^2,
\end{equation}
where $M =\frac{\beta}{\beta+\rho_1} U^k+\frac{\rho_1}{\beta+\rho_1} W^k$. 
Recall that $L(W)$ contains the cross-entropy, and there doesn't exist a closed-form solution. Thus, the stochastic gradient descent (SGD) optimizer is applied to each mini-batch. Let $t$ represent the number of iterations and $W^{k,0}=W^k$. It then derives 
\begin{equation}\label{SGD}
\begin{split}
   W^{k+1} &= W^{k,t-1} - \eta ( \nabla_W L(W^{k,t-1}) ) \\
   &\quad + (\beta + \rho_1) ( W^{k,t-1} - M )
\end{split}
\end{equation}
where $\nabla_W$ indicates the gradient of $L(W)$ at $W$ and $\eta$ is the learning rate. The overall
scheme is presented in Algorithm \ref{SGDalgo}.

\subsubsection{Update $U$}
The subproblem can be solved by
\begin{equation}\label{LU}
\begin{aligned}
    \min_U\quad \lambda\|U\|_{2,p}^{p}+\frac{\beta}{2}\|W^{k+1} -U\|_\textrm{F}^2+\frac{\rho_2}{2}\|U-U^k\|_\textrm{F}^2,
\end{aligned}
\end{equation}
which is equivalent to
\begin{equation}
    \min_U \quad\lambda\|U\|_{2,p}^{p}+\frac{\beta+\rho_2}{2}\|U-N\|_\textrm{F}^2,
\end{equation}
where $N=\frac{\beta}{\beta+\rho_2}W^{k+1}+\frac{\rho_2}{\beta+\rho_2}U^k$\vspace{1ex}.
It can be further decomposed into a series of channel optimization problems as 
\begin{equation}\label{U_i}
    \min_{U_{:j::}}\quad\lambda\|\textrm{vec}(U_{:j::})\|^p+\frac{\beta+\rho_2}{2}\|\textrm{vec}(U_{:j::})-\textrm{vec}(N_{:j::})\|^2,
\end{equation}
where $\textrm{vec}(U_{:j::})$ and $\textrm{vec}(N_{:j::})$ are the vectorized forms of $U_{:j::}$ and $N_{:j::}$, respectively. The proximal operator of $\| \cdot \|^p$ is reviewed in the following lemma.

\begin{lemma}\label{lemma2}
    Consider the proximal operator 
\begin{equation}\label{Prox}
\begin{aligned}
   \operatorname{Prox}_{{\lambda}|\cdot|^p}(a)&=\min_{x\in\mathbb{R}}{\lambda}|x|^p+\frac{1}{2}(x-a)^2\\
   &=\begin{cases}
       \{0\}, &|a|<\kappa({\lambda},p),\\
       \{0,\operatorname{sgn}(a)c({\lambda},p)\}, & |a|=\kappa({\lambda},p),\\
       \{\operatorname{sgn}(a)\varpi_p(|a|)\}, & |a|>\kappa({\lambda},p),
   \end{cases}
\end{aligned}    
\end{equation}
where
\begin{equation}
\begin{aligned}
c({\lambda}, p)&=(2 {\lambda}(1-p))^{\frac{1}{2-p}}>0,\\
\kappa({\lambda}, p)&=(2-p) {{\lambda}}^{\frac{1}{2-p}}(2(1-p))^{\frac{p+1}{p-2}},\\
\varpi_p(a) &\in \{x \mid x-a+{\lambda} p \operatorname{sgn}(x) x^{q-1}=0, x>0 \}.
\end{aligned}
\end{equation}
See \cite{zhou2023revisiting} for details.
\end{lemma}

Therefore, the solution of \eqref{U_i} is 
\begin{equation}\label{L_U}
\textrm{vec}(U_{:j::}^{k+1})\in
\operatorname{Prox}_{\frac{{\lambda}}{\rho_2+\beta}|\cdot|^p}
\!\left(\| \textrm{vec}(N_{:j::})\|\right)
\frac{N_j}{\|N_j\|}
\end{equation}
if $\| \textrm{vec}(N_{:j::})\|\neq0$ otherwise $\textrm{vec}(U_{:j::})=0$. Afterwards, $U^{k+1}$ can be obtained by rescaling $\textrm{vec}(U_{:j::}^{k+1})$ into the four-dimensional tensor.

\section{Experiments}

This section first compares the performance of our proposed GoPrune and ADMM \cite{ji2025network}, then visualizes the compression weight distribution, and finally discusses the effects of $p$ for GoPrune. All experiments are implemented with PyTorch 2.8 and run on a NVIDIA GeForce RTX 4090D.

\subsection{Setups}\label{setup}
In the experiments, some ResNet and VGG models are used, including ResNet-20\footnote{https://github.com/akamaster\label{web-resnet}}, ResNet-50\footnote{https://docs.pytorch.org/vision/stable/models/resnet.html}, ResNet-56\textsuperscript{\ref{web-resnet}}, ResNet-110\textsuperscript{\ref{web-resnet}}, VGG-16\footnote{https://docs.pytorch.org/vision/stable/models/vgg.html \label{web-vgg}}\, and VGG-19\textsuperscript{\ref{web-vgg}}. The selected evaluation datasets are CIFAR-10\footnote{https://www.cs.toronto.edu/~kriz/cifar\label{web-cifar}} and CIFAR-100\textsuperscript{\ref{web-cifar}}. 

In the compression stage, we set $\rho_1 = \rho_2 = 1.5 \times 10^{-3}$, $\beta = 1.5 \times 10^{-3}$, and $\alpha = 1\times 10^{-4}$. The weights are compressed towards zero through the PAM over 15 epochs. As for the value of $p$, for ADMM, $p = 1/5$, and for GoPrune, $p \in \{ 0, 1/2, 2/3 \}$. Let $I_{[c_j]_l}$ represent the importance score of the $c_j$-th channel in the $l$-th convolutional layer, which is 
\begin{equation}
    I_{[c_j]_l}=\frac{|W_{:j::}|-\min(|W_{:j::}|)}{\max(|W_{:j::}|)-\min(|W_{:j::}|)},
\end{equation}
where GoPrune performs channel pruning based on this importance score. All models are pruned to a ratio of 0.7, and fine-tuned for 300 epochs.  

In addition, two key metrics are considered, i.e., accuracy (ACC) and compression time. To ensure fairness and eliminate differences caused by different initial conditions, the experiments are repeated 35 times, and the results are presented as the mean and standard deviation.

\subsection{Experimental Results}\label{ACC-time}

Table \ref{CIFAR-10} and Table \ref{CIFAR-100} list the ACC results (mean$\pm$std) and compression time (in seconds) of ADMM  for unstructured pruning and GoPrune for structured pruning. 

It can be concluded that on the CIFAR-10 dataset, GoPrune is comparable to ADMM in terms of accuracy, and even shows some improvement. In terms of compression time, GoPrune significantly outperforms ADMM. Specifically, the average compression time is approximately 1.7 times faster than ADMM, with a compression time speedup of more than 2 times on ResNet-110  and 1.8 times on ResNet-56. This can be attributed to the larger number of parameters and their residual block architectures. Similar conclusions can be drawn on the CIFAR-100 dataset, where the average compression time is improved by 1.3 times.

\textbf{Overall, compared to the existing ADMM, our GoPrune not only maintains high accuracy and stability but also achieves a significant improvement in compression time.}

\begin{table}[t]
    \centering
    \caption{ACC results and compression time (in seconds) on the CIFAR-10 dataset.}\label{CIFAR-10}
     \scalebox{0.95}{{\renewcommand\baselinestretch{1.5}\selectfont
    \begin{tabular}{|c|c|c|c|c|}
        \hline
            \multirow{2}{*}{Models} &\multicolumn{2}{c|}{TOP-1 ACC (mean$\pm$std)}&\multicolumn{2}{c|}{Compression Time}\\
        \cline{2-5}
            &ADMM&GoPrune&ADMM&GoPrune \\
        \hline
        \hline   
           ResNet-20&88.83\%$\pm$0.10\%&\textbf{89.02\%$\pm$0.01\%} &172.56&\textbf{110.76} \\
        \hline
            ResNet-50&92.14\%$\pm$0.14\%&\textbf{92.69\%$\pm$0.13\%} &651.00&\textbf{435.18} \\
        \hline
            ResNet-56&90.07\%$\pm$0.14\%&\textbf{90.23\%$\pm$0.11\%} &392.58&\textbf{219.78} \\
        \hline
           ResNet-110&90.17\%$\pm$0.12\%&\textbf{90.56\%$\pm$0.07\%} &727.80&\textbf{358.20} \\
        \hline
            VGG-16&91.48\%$\pm$0.05\%&\textbf{91.87\%$\pm$0.10\%}&168.78&\textbf{108.60} \\  
        \hline
            VGG-19&91.24\%$\pm$0.05\%&\textbf{91.29\%$\pm$0.04\%}&213.40&\textbf{175.38} \\
        \hline
    \end{tabular}\par}}
\end{table}

\begin{table}[t]
    \centering
    \caption{ACC results and compression time (in seconds) on the CIFAR-100 dataset.}\label{CIFAR-100}
     \scalebox{0.95}{{\renewcommand\baselinestretch{1.5}\selectfont
    \begin{tabular}{|c|c|c|c|c|}
        \hline
            \multirow{2}{*}{Models} &\multicolumn{2}{c|}{TOP-1 ACC (mean$\pm$std)}&\multicolumn{2}{c|}{Compression Time}\\
        \cline{2-5}
            &ADMM&GoPrune&ADMM&GoPrune \\
        \hline
        \hline    
           ResNet-20&60.89\%$\pm$0.07\%&\textbf{61.10\%$\pm$0.05\%}&221.40&\textbf{145.78} \\
        \hline
            ResNet-50&70.80\%$\pm$0.08\%&\textbf{70.84\%$\pm$0.10\%}&654.00&\textbf{633.00} \\
        \hline
            ResNet-56&64.54\%$\pm$0.10\%&\textbf{64.86\%$\pm$0.05\%}&413.70&\textbf{294.00} \\
        \hline
           ResNet-110&65.35\%$\pm$0.46\%&\textbf{65.61\%$\pm$0.15\%}&693.60&\textbf{502.56} \\
        \hline
            VGG-16&69.24\%$\pm$0.09\%&\textbf{69.88\%$\pm$0.07\%}&164.22&\textbf{111.78} \\ 
        \hline
            VGG-19&67.53\%$\pm$0.07\%&\textbf{67.75\%$\pm$0.10\%}&198.00&\textbf{186.00} \\
        \hline
    \end{tabular}\par}}
\end{table}

\begin{figure*}[t]
    \centering
    \subfloat[ADMM on ResNet-50]{\includegraphics[width=0.24\linewidth]{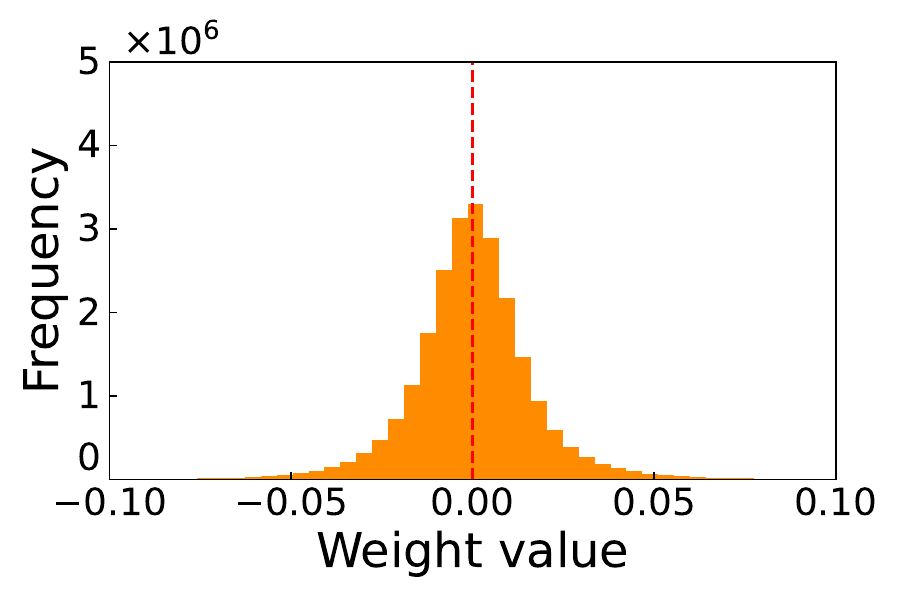}}
    \subfloat[ADMM on ResNet-110]{\includegraphics[width=0.24\linewidth]{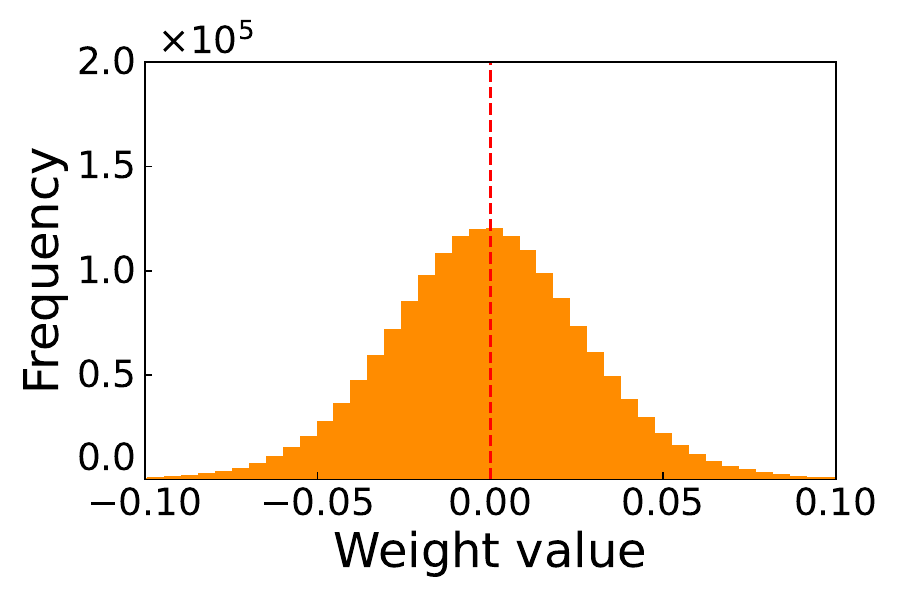}}
    \subfloat[ADMM on VGG-16]{\includegraphics[width=0.24\linewidth]{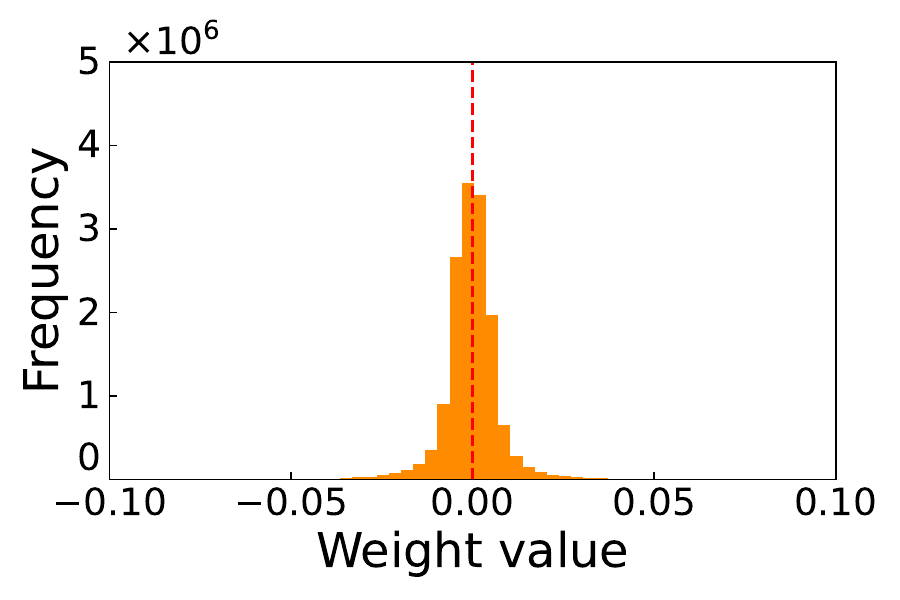}}
    \subfloat[ADMM on VGG-19]{\includegraphics[width=0.24\linewidth]{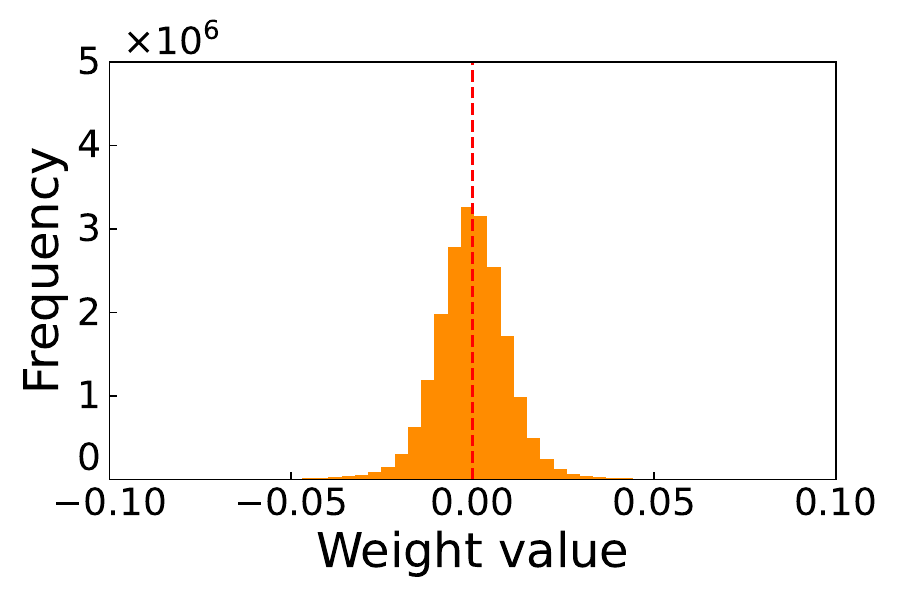}}
    
    \subfloat[GoPrune on ResNet-50]{\includegraphics[width=0.24\linewidth]{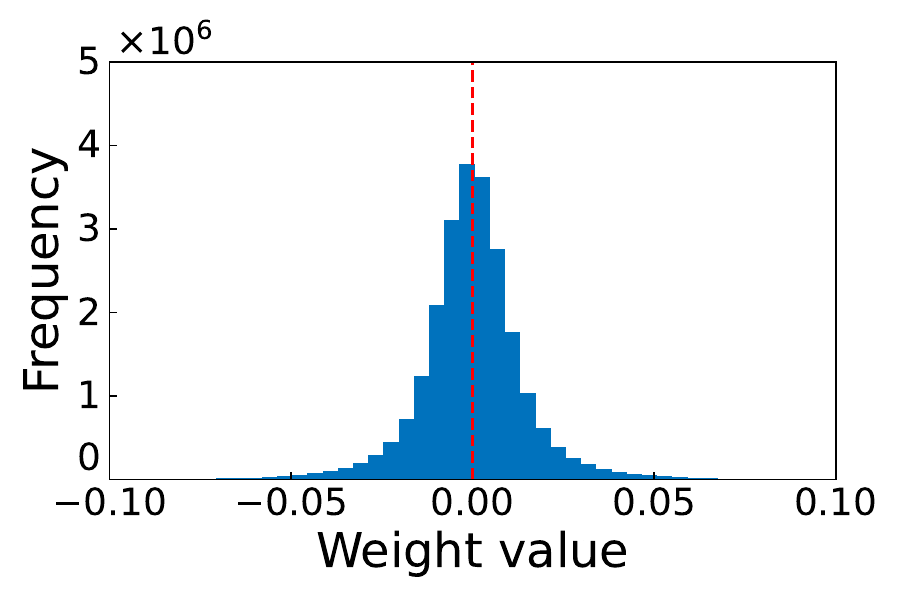}}
    \subfloat[GoPrune on ResNet-110]{\includegraphics[width=0.24\linewidth]{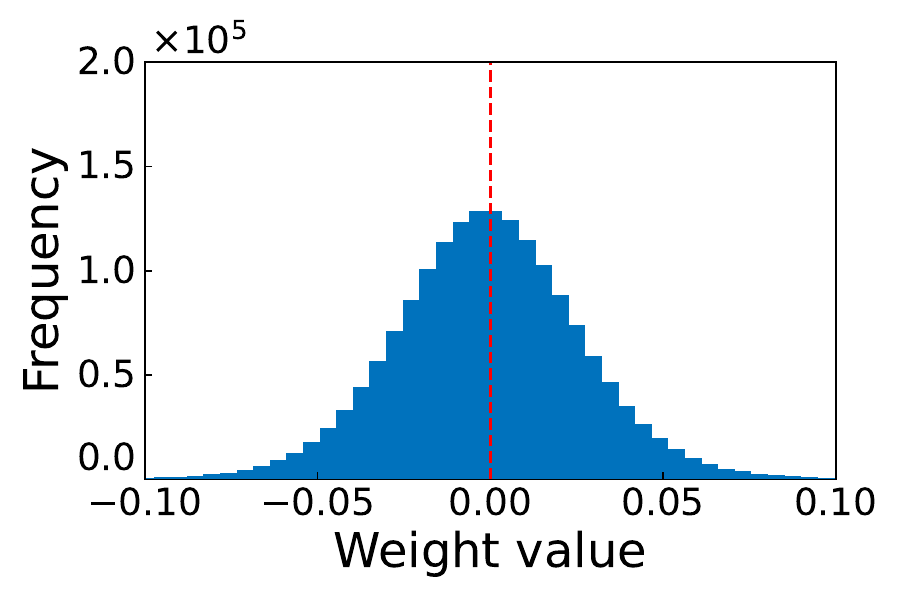}}
    \subfloat[GoPrune on VGG-16]{\includegraphics[width=0.24\linewidth]{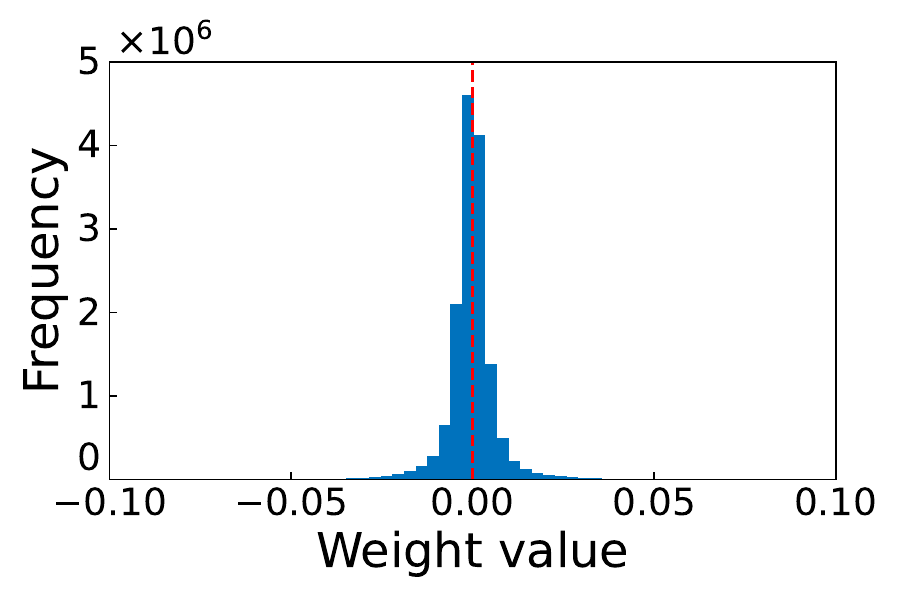}}
    \subfloat[GoPrune on VGG-19]{\includegraphics[width=0.24\linewidth]{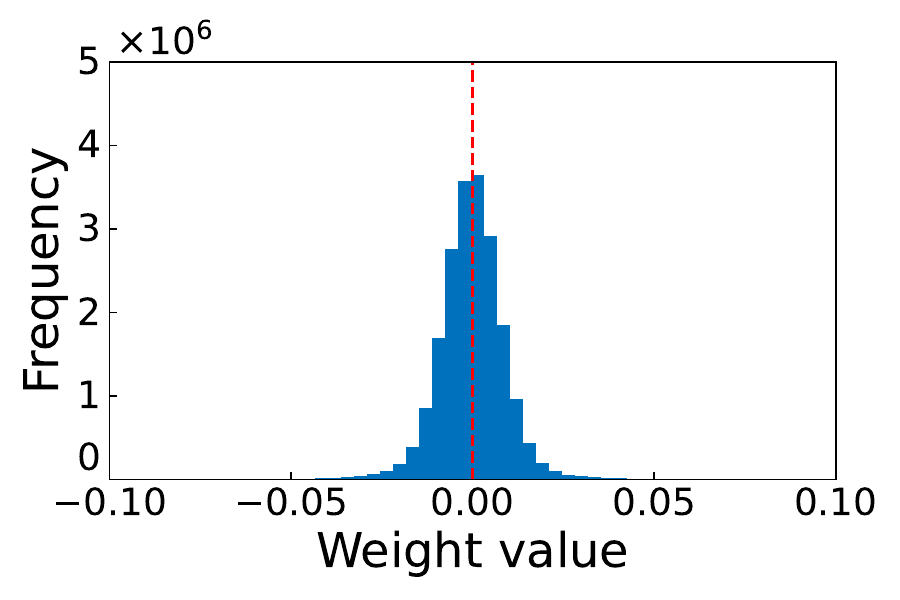}}
    
    \caption{Weights distribution on the CIFAR-10 dataset.}
    \label{wights com10}
\end{figure*}

\begin{figure*}[t]
    \centering
     \subfloat[ADMM on ResNet-50]{\includegraphics[width=0.24\linewidth]{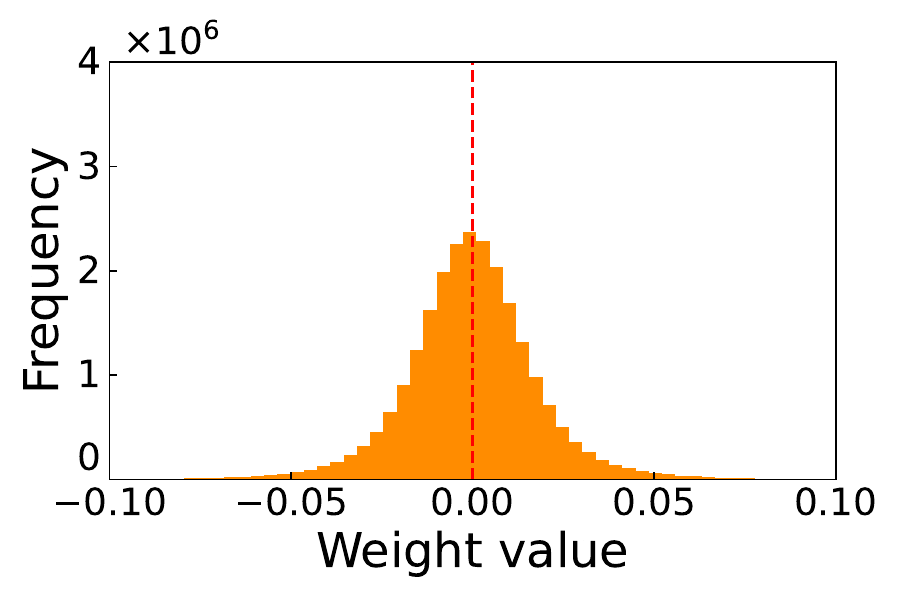}}
     \subfloat[ADMM on ResNet-110]{\includegraphics[width=0.24\linewidth]{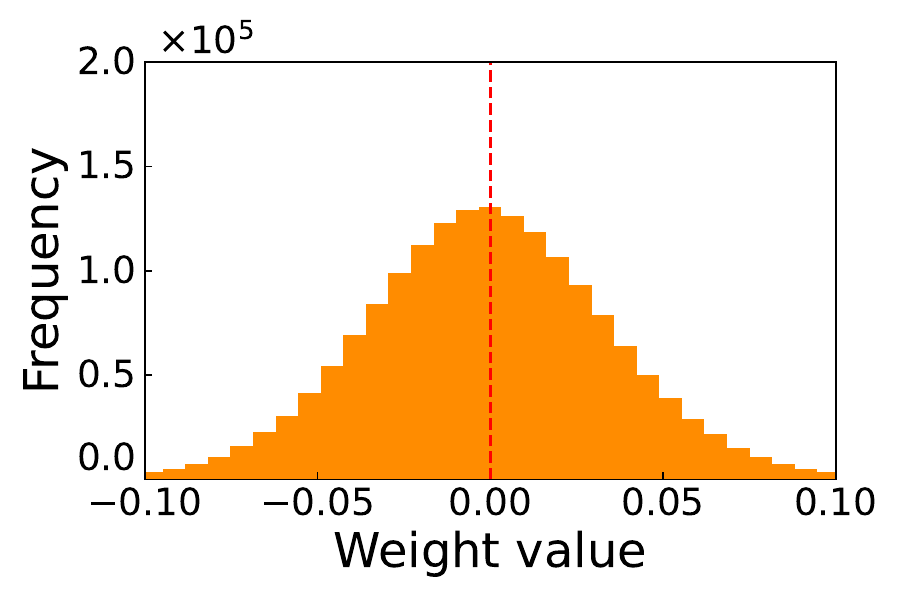}}
    \subfloat[ADMM on VGG-16]{\includegraphics[width=0.24\linewidth]{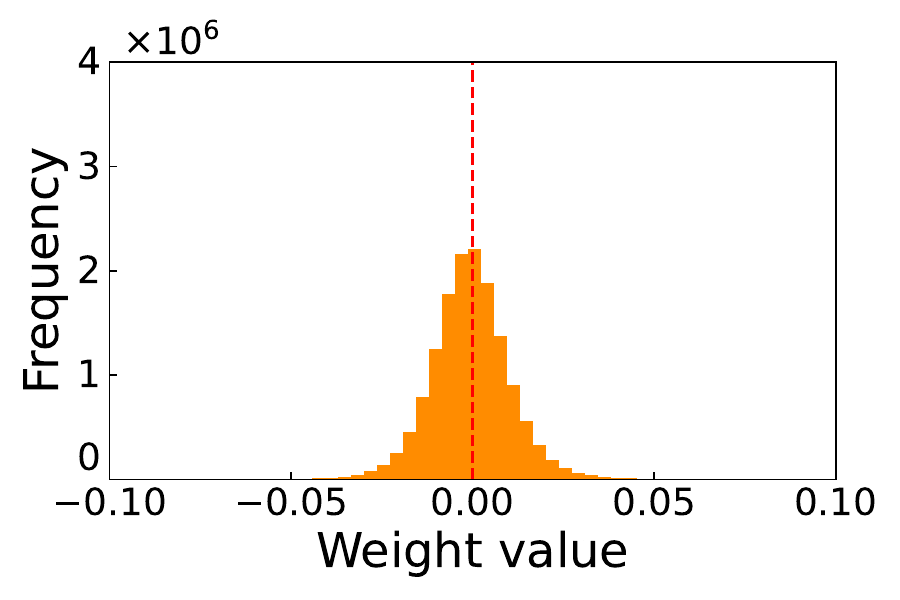}}
    \subfloat[ADMM on VGG-19]{\includegraphics[width=0.24\linewidth]{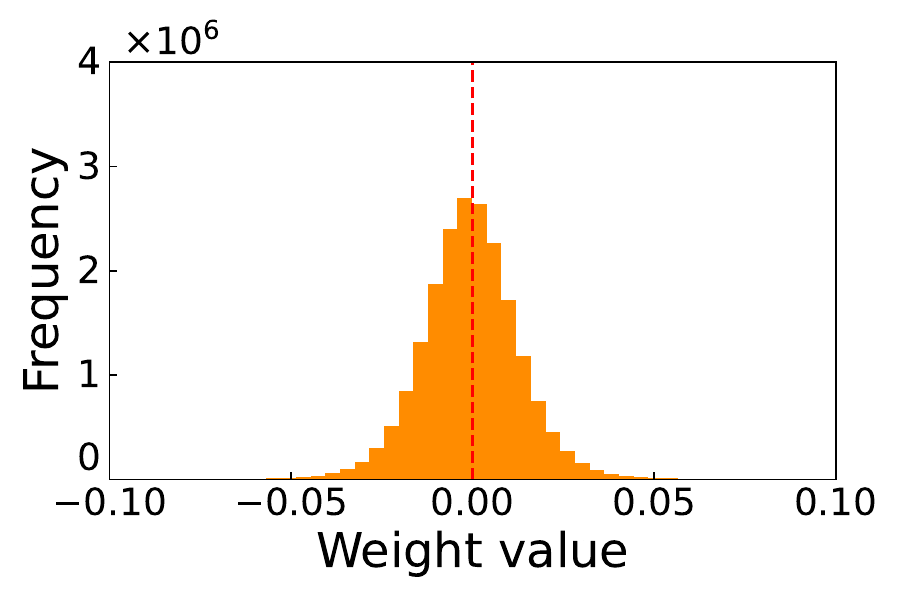}}

     \subfloat[GoPrune on ResNet-50]{\includegraphics[width=0.24\linewidth]{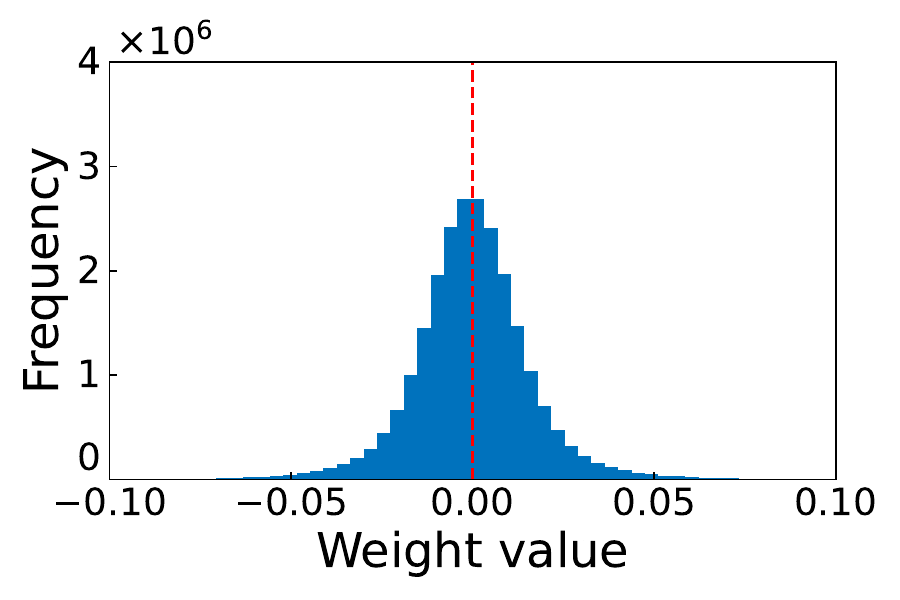}}
    \subfloat[GoPrune on ResNet-110]{\includegraphics[width=0.24\linewidth]{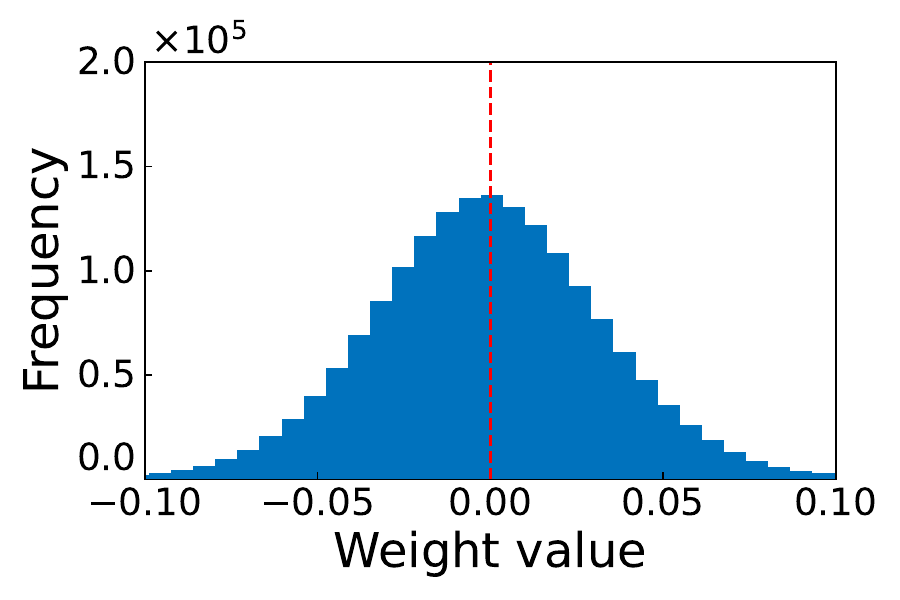}}
    \subfloat[GoPrune on VGG-16]{\includegraphics[width=0.24\linewidth]{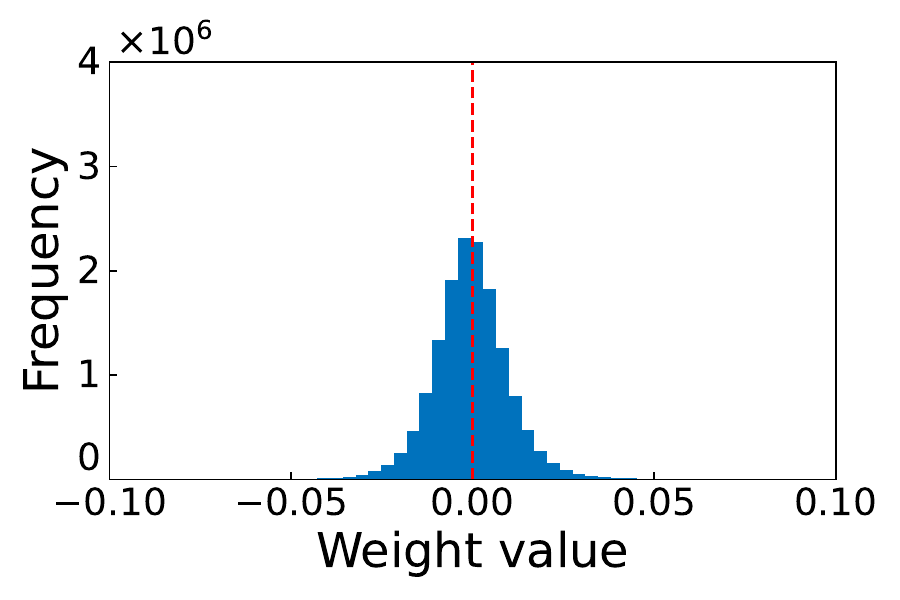}}
   \subfloat[GoPrune on VGG-19]{\includegraphics[width=0.24\linewidth]{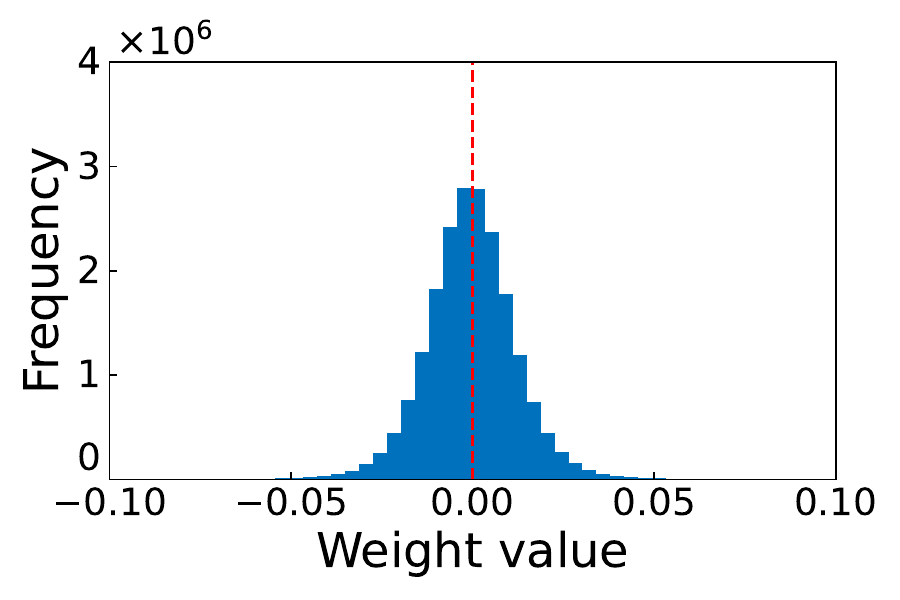}}
   
    \caption{Weights distribution on the CIFAR-100 dataset.}
    \label{wights com100}
\end{figure*}

\subsection{Weights Distribution}\label{weights}
This subsection investigates the weight-distribution evolution during the compression procedures. Fig. \ref{wights com10} and Fig. \ref{wights com100} visualize the compressed results of ResNet-50, ResNet-110, VGG-16, and VGG-19 on the CIFAR-10 and CIFAR-100 datasets, respectively. 
Here, yellow represents the weight distribution after ADMM compression, and blue represents the weight distribution after GoPrune compression. It is clear to see that our GoPrune is able to push more weights toward zero compared with ADMM, with the most significant improvement observed on VGG-16 on the CIFAR-10 dataset.

\textbf{Overall, combined with Subsection \ref{ACC-time}, our proposed GoPrune achieves excellent compression efficiency while maintaining high accuracy and less compression time.}

\subsection{Effects of $p$}\label{effect of p}
\begin{figure}[t]

    \centering
    \subfloat[ResNet-20 ]{
        \includegraphics[width=0.48\linewidth]{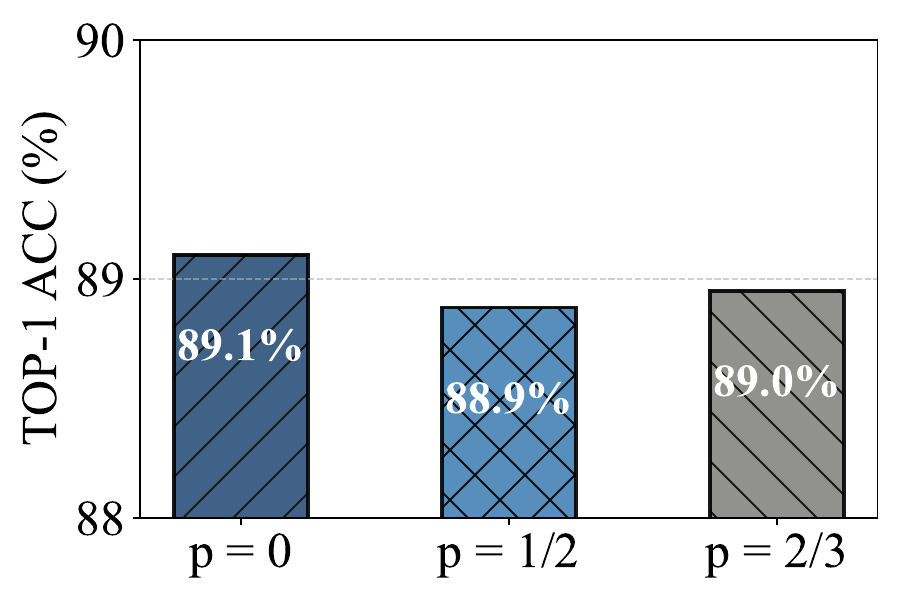}}
    \subfloat[ResNet-50 ]{
        \includegraphics[width=0.48\linewidth]{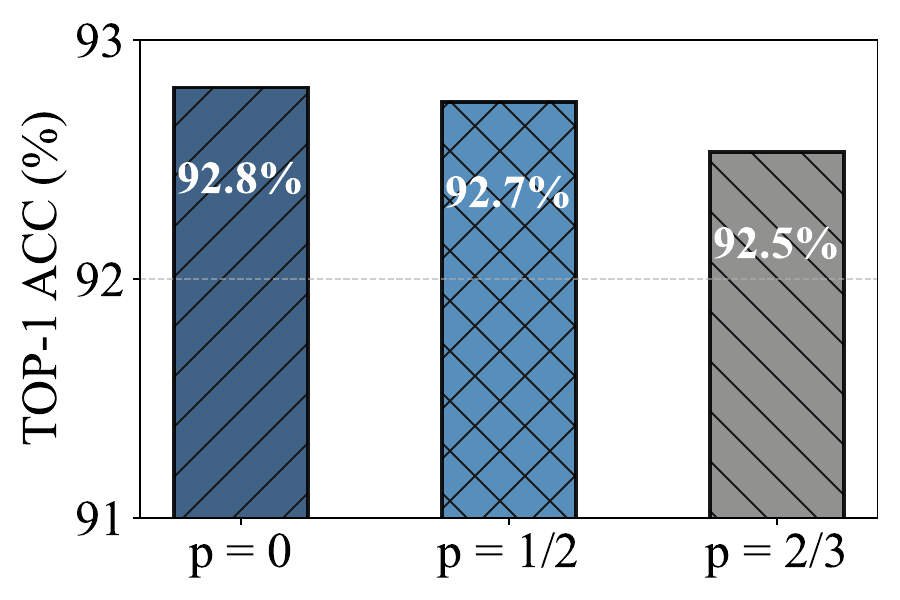}}
        
    \subfloat[ResNet-56 ]{
        \includegraphics[width=0.48\linewidth]{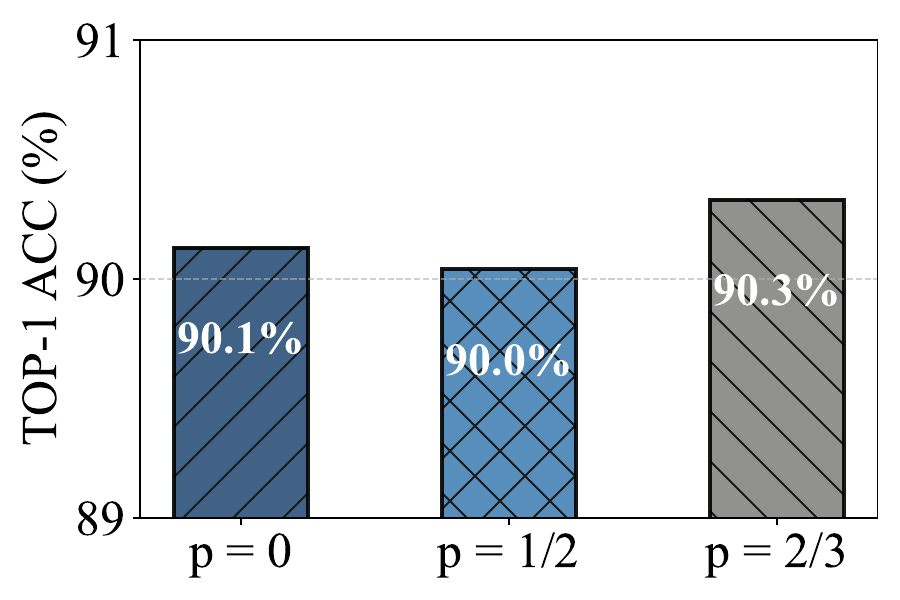}}
    \subfloat[ResNet-110 ]{
        \includegraphics[width=0.48\linewidth]{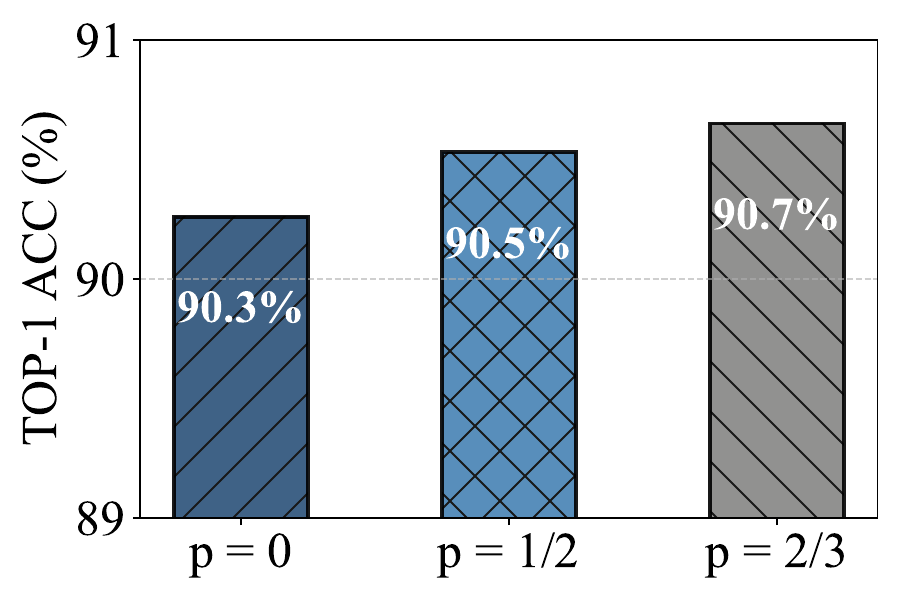}}    
        
    \subfloat[VGG-16  ]{
        \includegraphics[width=0.48\linewidth]{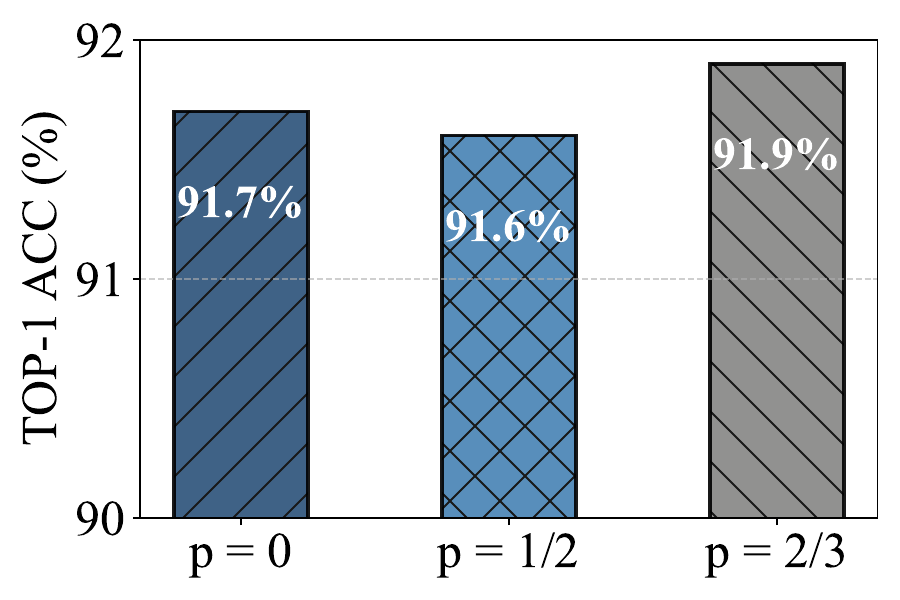}} 
    \subfloat[VGG-19 ]{
        \includegraphics[width=0.48\linewidth]{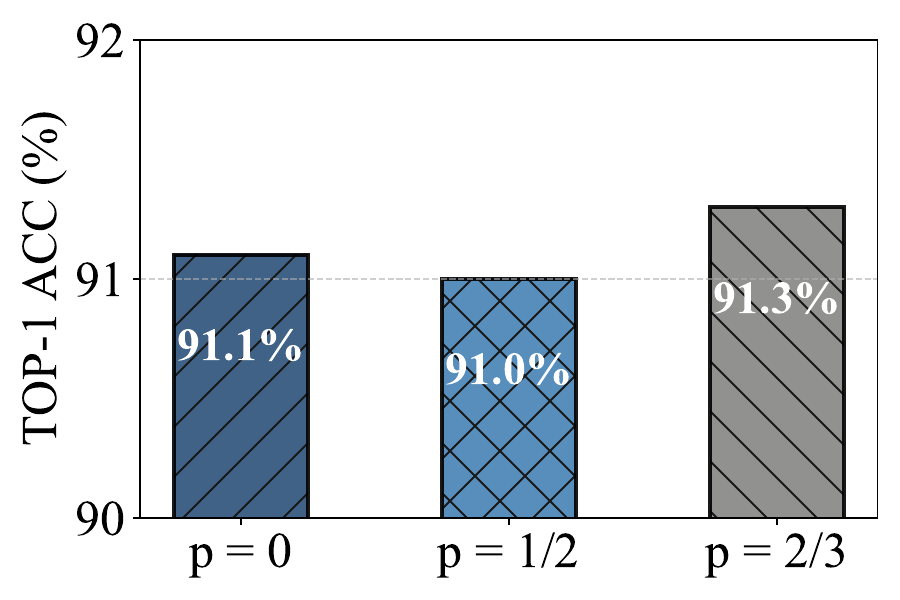}}    
       \caption{Effects of $p$ on the CIFAR-10 dataset.}
       \label{optimal p10}
\end{figure}

\begin{figure}[t]

    \centering
    \subfloat[ResNet-20 ]{
        \includegraphics[width=0.48\linewidth]{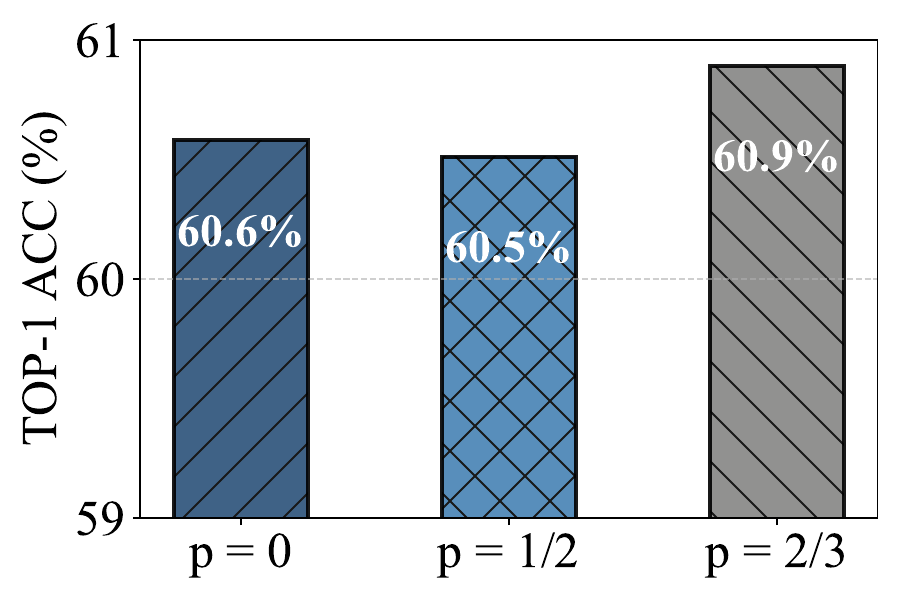}}
    \subfloat[ResNet-50 ]{
        \includegraphics[width=0.48\linewidth]{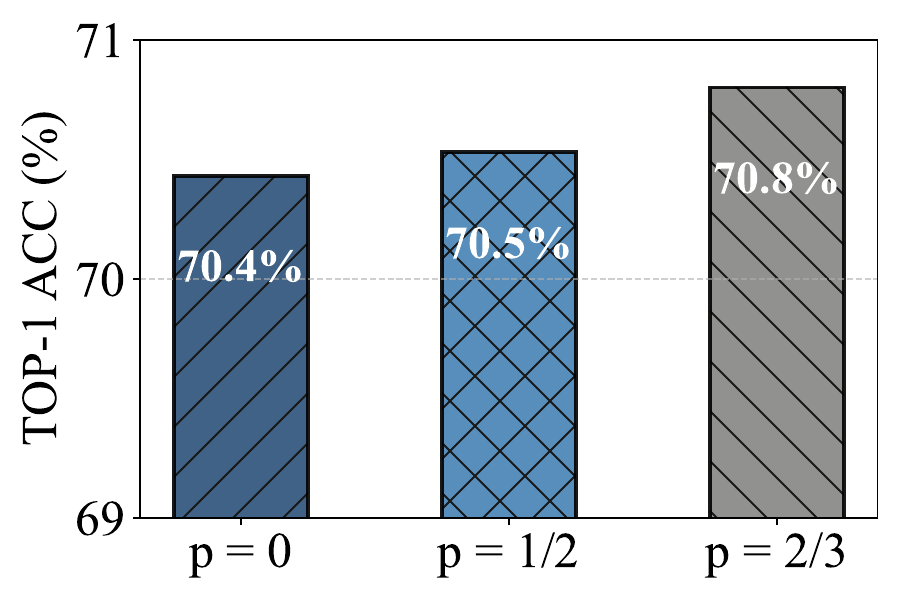}}
        
    \subfloat[ResNet-56 ]{
        \includegraphics[width=0.48\linewidth]{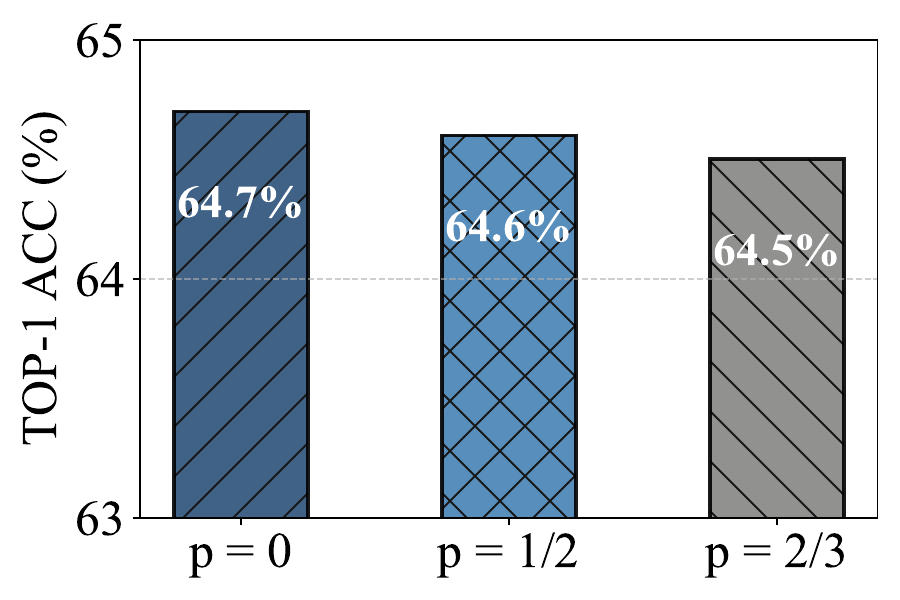}} 
    \subfloat[ResNet-110 ]{
        \includegraphics[width=0.48\linewidth]{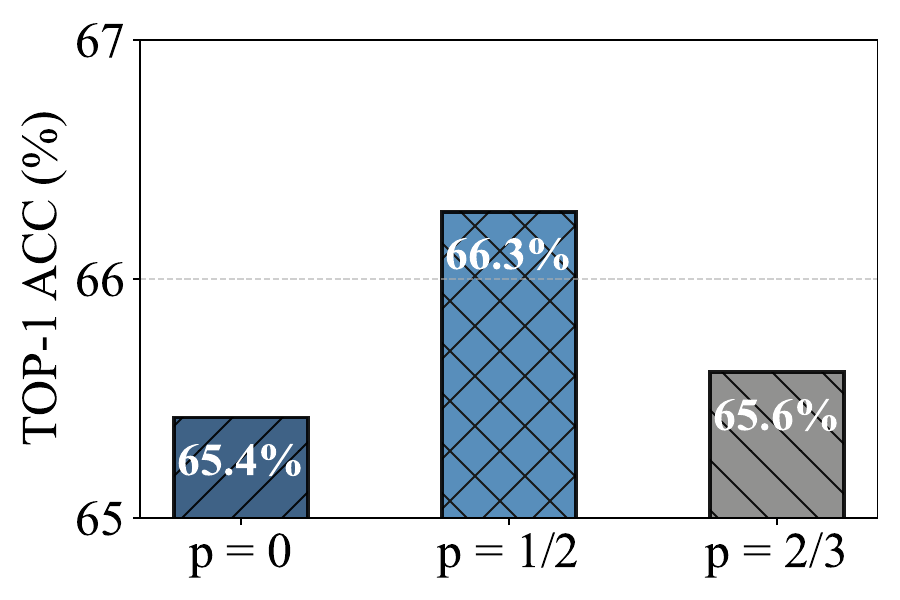}}    

    \subfloat[VGG-16 ]{
        \includegraphics[width=0.48\linewidth]{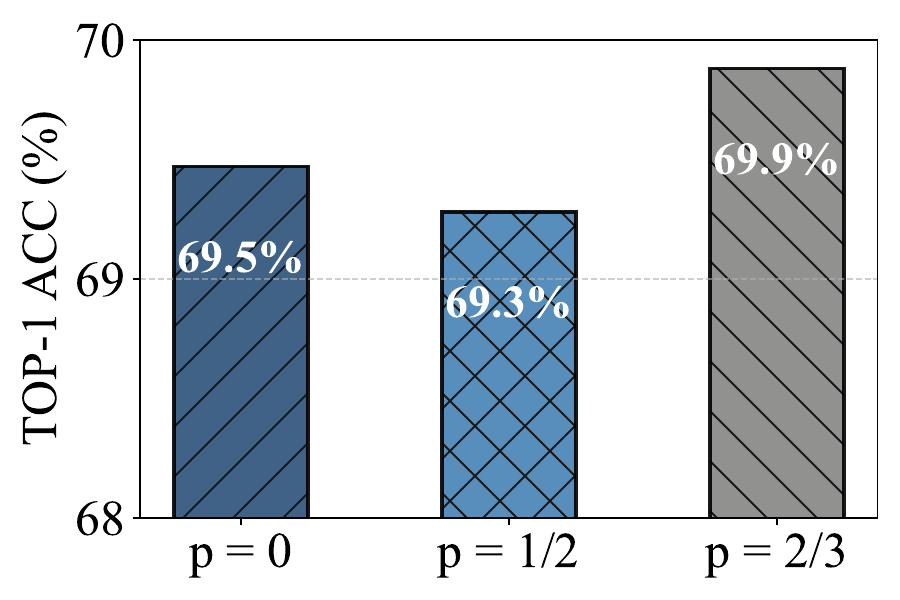}} 
    \subfloat[VGG-19 ]{
        \includegraphics[width=0.48\linewidth]{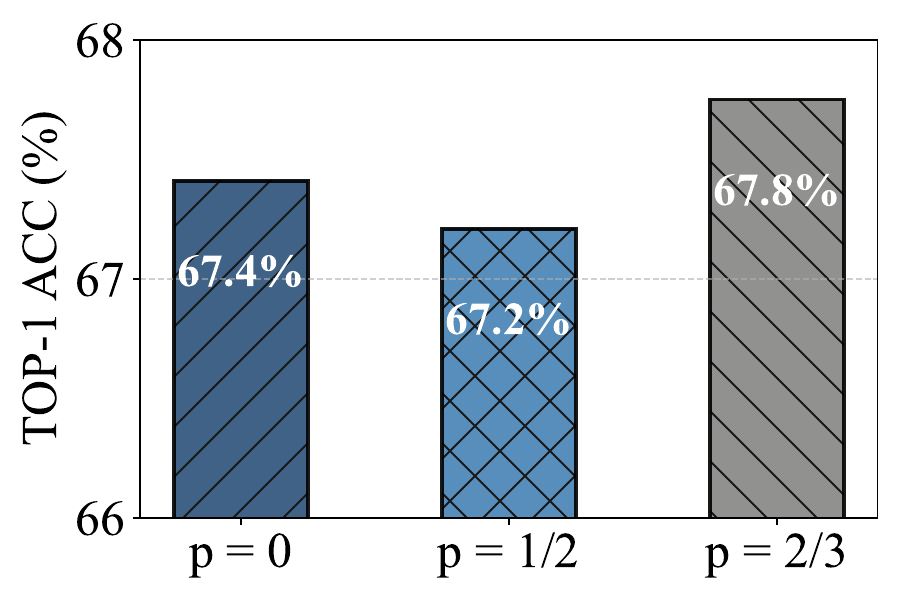}}    
       \caption{Effects of $p$ on the CIFAR-100 dataset.}
       \label{optimal p100}
\end{figure}

This subsection discusses the accuracy after compression, pruning, and fine-tuning under three different values of $p$. Fig. \ref{optimal p10} presents the experimental results on the CIFAR-10 dataset. It can be observed that for ResNet-20 and ResNet-50, the best accuracy is achieved when $p = 0$, while for the other models, the highest accuracy occurs at $p = 2/3$. Fig. \ref{optimal p100} shows the results on the CIFAR-100 dataset. It can be seen that ResNet-110 achieves the best accuracy when $p = 1/2$, ResNet-56 performs best when $p = 0$, and the remaining models achieve their highest accuracy at $p = 2/3$.

\textbf{Overall, the optimal value of $p$ varies across different models and architectures, which validates that  extending $p$ from $(0,1)$ to $[0,1)$ is of importance.}

\section{Conclusion}

This paper proposes a highly efficient structured pruning framework named GoPrune, which is based on the $\ell_{2,p}$-norm regularization with $p\in[0,1) $ and optimized via the PAM algorithm. Unlike the existing ADMM-based $\ell_{p}$-norm unstructured pruning that achieve sparsity only on individual weights, GoPrune calculates the $\ell_{2}$-norm within each channel and applies the $p$-norm at the channel level, thereby achieving channel-level structured sparsity. Numerical studies validate that our GoPrune outperforms the ADMM-based baseline in both pruning accuracy and compression efficiency. 

In the future, we are interested in  extending this method to prune large language models.

\bibliographystyle{ieeetr}
\bibliography{mybibfile}

\end{document}